\documentclass{svproc}
\usepackage{url}
\usepackage{graphicx}
\usepackage{adjustbox}

\begin{document}
\mainmatter              % start of a contribution
\title{TSEM: Temporally-Weighted Spatiotemporal Explainable Neural Network for Multivariate Time Series}
\titlerunning{TSEM}  % abbreviated title (for running head)
%                                     also used for the TOC unless
%                                     \toctitle is used
%
\author{Anh-Duy Pham\inst{1} \and Anastassia Kuestenmacher\inst{1}
 \and
Paul G Ploeger \inst{1}}
\authorrunning{Anh-Duy Pham et al.} % abbreviated author list (for running head)
%
%%%% list of authors for the TOC (use if author list has to be modified)
\tocauthor{Anh-Duy Pham, Anastassia Kuestenmacher and
Paul G Ploeger}
\institute{Hochschule Bonn-Rhein-Sieg\\
\email{{duy.pham}@smail.inf.h-brs.de, \\ \{anastassia.kuestenmacher, paul.ploeger\}@h-brs.de}}

\maketitle              % typeset the title of the contribution

\begin{abstract}
Deep learning has become a one-size-fits-all solution for technical and business domains thanks to its flexibility and adaptability. It is implemented using opaque models, which unfortunately undermines the outcome's trustworthiness. In order to have a better understanding of the behavior of a system, particularly one driven by time series, a look inside a deep learning model so-called post-hoc eXplainable Artificial Intelligence (XAI) approaches, is important. There are two major types of XAI for time series data: model-agnostic and model-specific. Model-specific approach is considered in this work. While other approaches employ either Class Activation Mapping (CAM) or Attention Mechanism, we merge the two strategies into a single system, simply called the Temporally Weighted Spatiotemporal Explainable Neural Network for Multivariate Time Series (TSEM). TSEM combines the capabilities of RNN and CNN models in such a way that RNN hidden units are employed as attention weights for the CNN feature maps' temporal axis. The result shows that TSEM outperforms XCM. It is similar to STAM in terms of accuracy, while also satisfying a number of interpretability criteria, including causality, fidelity, and spatiotemporality.
% We would like to encourage you to list your keywords within
% the abstract section using the \keywords{...} command.
\end{abstract}
\section{Introduction}
\begin{figure}[h]
\centering
\includegraphics{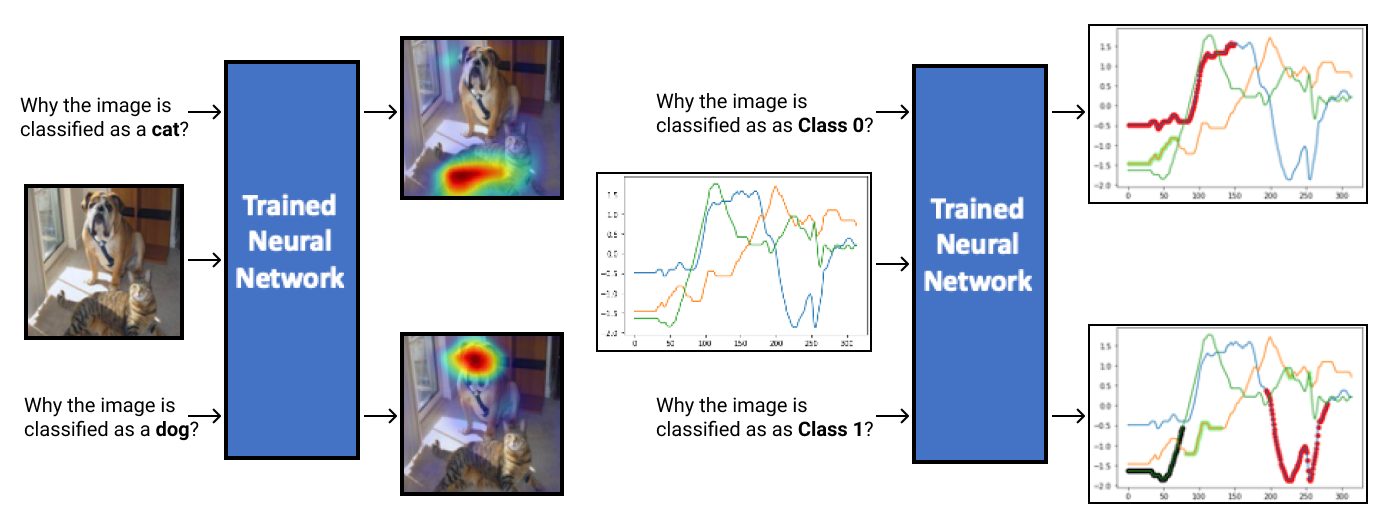}
\caption{Explanations given by Gradient-based Class Activation Mapping method}
\label{fig:XAI_gen}
\end{figure}

Multivariate time series analysis is used in many sensor-based industrial applications. Several advanced machine learning algorithms have achieved state-of-the-art classification accuracy in this field, but they are opaque because they encode important properties in fragmented, impenetrable intermediate layers. In Fig. \ref{fig:XAI_gen}, the positions of the cat and the dog are clear, but the signal from the three sensors is perplexing, even with explanations. This is perilous because adversarial attacks can exploit this confusion by manipulating the inputs with intentional noise to trick the classification model to yield a wrong decision. Multivariate time series (MTS) are challenging to classify due to the underlying multi-dimensional link among elemental properties. As Industry 4.0 develops, a sensor system is incorporated into production and operational systems to monitor processes and automate repetitive tasks. Hence, given the high accuracy of multivariate time series classification algorithms, it is crucial to understand the precise rationale for their conclusions, particularly in precision manufacturing or operations such as autonomous driving. This is when interpretable techniques become useful. 

Generally, in the diversity of interpretable AI methods, there are three main categories, namely interpretable-by-design, model-agnostic and model-specific ones. The interpretable-by-design models are those that were built with the intention of giving an explanation. Although they are of classical methods, such as decision trees and general linear models, they still give contributions to the novel interpretable applications as graph neural networks \cite{pfeifer2022gnn}, a decision tree \cite{nauta2021neural}, or by comparing to archetypal instances \cite{chen2019looks}, conditioning neurons to interpretable properties \cite{hu2019introductory,koh2020concept}, and statistically accounting evidence from various image patches \cite{brendel2019approximating}. On the other hand, model-agnostic and model-specific explainable methods are applied to models that are not interpretable-by-design. The model-agnostic techniques are methods that may be used to any predictive model, while the model-specific methods are methods that are linked to a particular model class. In this work, only model-specific interpretable methods are analyzed. 

Post-hoc analysis, which explains the decision of a model by analyzing its output, and ante-hoc analysis, which provides an explanation as part of its decision, are the two ways that explanations could be extracted from a model regardless of whether the method is model-agnostic or model-specific. Nevertheless, there are two ways that explanations could be extracted from a model regardless of whether the method is model-specific or model-agnostic \cite{holzinger2022xxai}. Class Activation Mapping (CAM)-based approaches that give post-hoc explanations for the output of a Convolutional Neural Network (CNN) and Attention-based Recurrent Neural Networks (RNNs) with their attention vectors as the ante-hoc explanation are analyzed in this research. 

This study proposes the Temporally weighted Spatiotemporal Explainable Network for Multivariate Time Series (TSEM), a novel method that overcomes several shortcomings of previous interpretations by leveraging the power of RNN to extract global temporal features from the input as an importance weight vector for the CNN-generated spatiotemporal feature maps in a parallel branch of the network. We show that our method could fix the locality of the explanations along the temporal axis yielded by 1D convolutional layers, while also improving the overall performance of the model, as the temporal information could now be directly weighted into the feature maps, as opposed to merely serving as supplementary information, as it is the case in XCM \cite{fauvel2021xcm}. Then, the saliency maps are extracted from the weighted feature maps by CAM methods. Despite the fact that CAM only gives local fidelity with CNN saliency maps, attention neural models are unable to provide consistent explanations for classification RNN models when compared to CAM explanations \cite{jain2019attention}. Apart from being faithful, such an explanation should satisfy two additional assessment criteria, namely spatiotemporal explainability, and causality \cite{gangopadhyay2021spatiotemporal}.

The rest of this paper is arranged in the following manner. Section 2 summarizes recent research on transparency in the realm of MTS classification. Section 3 explains the novel TSEM architecture and how its outputs should be interpreted. Section 4 details the experiments and methods of assessment. Section 5 summarizes the method's accomplishments and offers more perspectives on how to improve it. 
\section{Related Work}
\label{sec:rel}
\subsection{Attention neural models}
\label{sec:rel1}
In terms of efficiency in learning long-term correlations between values in a time series, the RNN layer may also be preserved in a neural network to retain information over an extended period of time, enabling more precise classification. The interpretation is then compensated by wrapping an attention neural model around the RNN layer to obtain additional information about the time series region of interest, which may improve the learning operation of the RNN layer. In addition, the attention neural model may offer input to the coupling RNN layer, instructing it to highlight the most important data.

Numerous MTS classification and regression models have been published. This kind of architectures begins with the Reverse Time Attention Model (RETAIN) \cite{choi2016retain}, which utilizes a two-layer neural attention model in reverse time order to emphasize the most recent time steps. Dual-Stage Attention-Based Recurrent Neural Network (DA-RNN) \cite{qin2017dual} adopts the concept of RETAIN but without the temporal order reversal, while Dual-Stage Two-Phase attention-based Recurrent Neural Network (DSTP-RNN) \cite{liu2020dstp} advances DA-RNN by dividing the spatial attention stage into two substages, or two phases. The Multi-level attention networks for geo-sensory time series (GeoMAN) \cite{liang2018geoman}similarly has two stages of attention, but in the spatial stage, it incorporates two parallel states: the local one correlates intra-sensory interactions, while the global one correlates inter-sensory ones. Spatiotemporal attention for multivariate time series prediction (STAM) \cite{gangopadhyay2021spatiotemporal} focused primarily on temporal feature extraction by assigning two attention vectors with independent RNN wrappers while decreasing the number of phases in the spatial attention stage to one. 

\subsection{Post-hoc model-specific convolutional neural network -based models}
\label{sec:rel2}
An LSTM layer is designed to learn the intercorrelation between values along the time dimension, which is always a one-dimensional sequence; hence, the layer can be replaced solely by a one-dimensional convolutional layer. Using this concept, Roy Assaf et al. \cite{assaf2019mtex} created the Multivariate Time Sequence Explainable Convolutional Neural Network (MTEX-CNN), a serial two-stage convolutional neural network consisting of a series of Conv2D layers coupled to a series of 1D convolutional (Conv1D) layers. Similarly, Kevin Fauvel et al. \cite{fauvel2021xcm} obtained the two convolutional layers by linking them in parallel, believing that this would give an extra temporal interpretation for the neural network's predictions. The eXplainable Convolutional Neural Network for Multivariate Time Series Classification (XCM) architecture provides a faithful local explanation to the prediction of the model as well as a high overall accuracy for the prediction of the model. Furthermore, as shown by the author of this research, the CNN-based structure allows the model to converge quicker than the RNN-based structure while also having lesser variation across epochs. Besides that, in their paper \cite{fauvel2021xcm}, XCM demonstrated their state of the art performance. By switching from a serial to a parallel framework, XCM was able to considerably enhance the performance of MTEX-CNN. Specifically, the information from the input data is received directly by both 1D and 2D modules; as a result, the extracted features are more glued to the input than if the data were taken from another module. 

However, they combine these two types of information, known as spatial and temporal features, by concatenating the temporal one as an additional feature vector to the spatial feature map and then using another Conv1D layer to learn the association between the temporal map and the spatial map, as shown in Fig. \ref{fig:XCM_arch}. This approach entails the following limitation: (1) The intermediate feature map has the form (feature length + 1, time length), and it leads to the explanation map, which has the same size as the intermediate feature map. This is out of sync with the size of the input data, which is (feature length, time length) and (2) while it is assumed that the final Conv1D layer will be able to associate the relationship between the concatenated map, the Conv1D layer will only be able to extract local features, whereas temporal features require long-term dependencies in order to achieve a more accurate correlation between time steps. 

In order to address these issues, TSEM is proposed as an architecture that makes better use of the temporal features by adding up values to the spatial features, even seemingly spatiotemporal features, yielded from the 2D convolutional (Conv2D) layers by multiplying them together and replaces the Conv1D layer with an LSTM layer, which can be better weighted in the spatial-temporal feature maps.

\subsection{Explanation extraction by Class Activation Mapping}
\label{sec:rel3}
Class Activation Mapping (CAM) methods determine which input characteristics are accountable for a certain categorization result. This is accomplished by performing backpropagation from the output logits to the desired layer to extract the activation or saliency maps of the appropriate features, and then interpolating these maps to the input to emphasize the accountable ones. CAM \cite{zhou2016learning} is the original approach, which uses the max-pooling layer to link the final convolutional layer with the logits layer in order to immediately remedy the liable features in the latter. Then, CAM becomes the general method name for this strategy and is further classified into two domains, excluding itself: gradient-based CAM and score-based CAM. The gradient-based CAM group consists of algorithms that backpropagate the gradient from the logits layer in order to weight the feature maps of the associated convolutional layer. They are listed as Grad-CAM \cite{selvaraju2017grad}, Grad-CAM++ \cite{chattopadhay2018grad}, Smooth Grad-CAM++ \cite{omeiza2019smooth}, XGrad-CAM \cite{fu2020axiom}, and Ablation-CAM \cite{ramaswamy2020ablation}, and are distinguished by their formulation for the combination of the backpropagated weights gradients and the weight. In contrast, score-based approaches such as Score-CAM \cite{wang2020score}, Integrated Score-CAM \cite{naidu2020cam}, Activation-Smoothed Score-CAM \cite{wang2020ss}, and Input-Smoothed Score-CAM \cite{wang2020ss} employ the logits to directly weight the convolutional layer of interest. They also vary in their concept of the combination of scores and feature maps through multiplication.
\section{Methodology}
The XCM acquires a basic CNN developed to extract features from the input data's variables and timestamps. It ensures the model choice made using Grad-CAM \cite{selvaraju2017grad} is interpretable faithfully. On a variety of UEA datasets, XCM beats state-of-the-art techniques for MTS classification \cite{bagnall2018uea}. Since faithfulness evaluates the relationship between the explanation and what the model computes, it is critical when describing a model to its end-user. The purpose of this study is to develop a small yet scalable and explainable CNN model that is true to its prediction.
The combination of CNN architecture with Grad-CAM enables the creation of designs with few parameters while maintaining accuracy and transparency. MTEX-CNN demonstrated the preceding by proposing a serial connection of 2D and Conv1D layers for the purpose of extracting essential characteristics from MTS.  
\begin{figure}[h]
\centering
\includegraphics[width=1.2\textwidth]{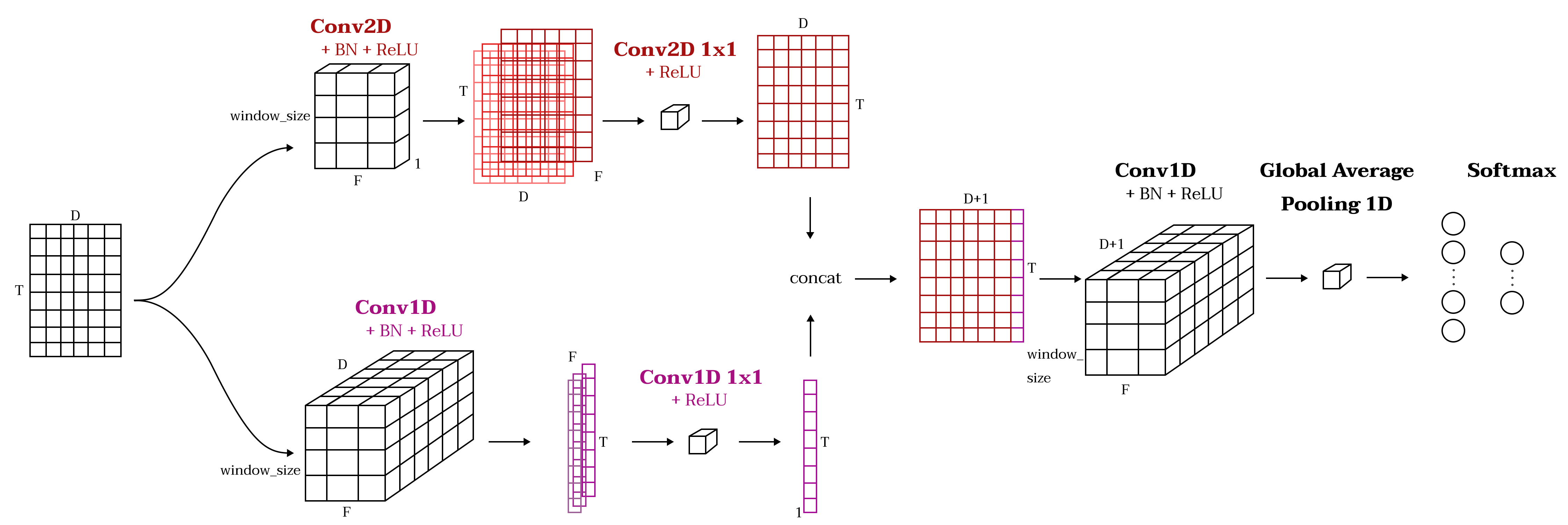}
\caption{XCM architecture \protect\cite{fauvel2021xcm}. Abbreviations: BN—Batch Normalization, D—number of observed variables,
F—number of filters, T—time series length and Window Size—kernel size, which corresponds
to the time window size.}
\label{fig:XCM_arch}
\end{figure}

\begin{figure}[h!]
\centering
\includegraphics[width=1.2\textwidth]{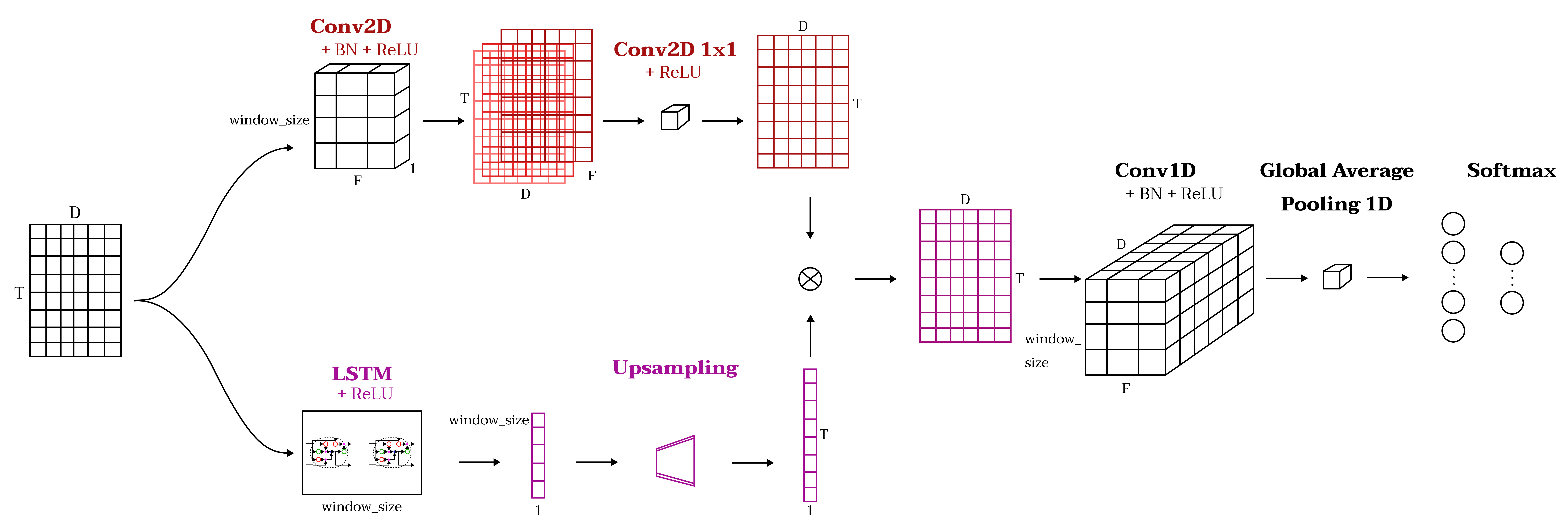}
\caption{TSEM architecture. Abbreviations: BN—Batch Normalization, D—number of observed variables,
F—number of filters, T—time series length and Window Size—kernel size, which corresponds
to the time window size.}
\label{fig:TSEM_arch}
\end{figure}

To leverage the above mentioned drawback of CNN post-hoc explanations, TSEM takes the backbone of the XCM architecture and improves it by replacing the Conv1D module, which includes two Conv1D layers in the second parallel branch of the architecture, with a single recurrent layer in the first parallel branch of the architecture, as previously stated. The time window aspect of the model has also been retained since it aids in scaling the model to a fraction of the input size when the data dimensions are too huge. Fig. \ref{fig:TSEM_arch} depicts the overall architecture. 

Formally, the input MTS is simultaneously fed into two different branches that consider each of its two dimensions, namely spatial and temporal ones. The spatial branch is designed to extract the spatial information spanned across the constituent time series by firstly applying a convolutional filter with a customized kernel size with one side fixed to the length of temporal axis of the MTS. This is done to reduce the number of model parameters and increase the training as well as inference speed. It is then followed by a 1x1 Conv2D layers to collapse the number of previous filters into one filter, making the shape of the resulting feature map equalled to the input. The idea is proposed by Fauvel et al. \cite{fauvel2021xcm} in their XCM architecture and kept as the backbone in our architecture, TSEM.

In the temporal branch, since the temporal explanation is redundant when the first branch can extract the spatiotemporal explanation, the Conv1D module is replaced with an LSTM module with a number of hidden units equal to the window size hyperparameter. The substitution sacrifices the explainability of the Conv1D module in exchange for improved temporal features because the LSTM module treats the time series signal as a continuous input rather than a discrete one as in the Conv . It is then upsampled to the size of the original time duration and element-wise multiplied with the feature maps from the first branch's two-dimensional convolutional module instead of being concatenated as in the temporal branch of XCM (Fig. \ref{fig:XCM_arch}). This results in time-weighted spatiotemporal forward feature maps from which the explanation map may be retrieved by different CAM-based techniques. Additionally, the new feature map is considered to improve accuracy when compared to XCM.
\section{Experiments and evaluation}
The assessment entails conducting tests for critical metrics that an interpretation should adhere to, as recommended and experimented with in several works on interpretable or post-hoc interpretability approaches.
\subsection{Baselines}

TSEM is evaluated in comparison to all of the attention neural models as well as post-hoc model-specific CNN-based models that are outlined in Sections \ref{sec:rel1}, \ref{sec:rel2}. These includes five attention neural models, namely RETAIN, DA-RNN, DSTP-RNN, GeoMAN and STAM as well as some of their possible variants along with MTEX-CNN and XCM, which are the two model-specific interpretable models. It is important to note that in the interpretability tests, the post-hoc analysis of MTEX-CNN and XCM is supplied by each of the explanation extraction techniques stated also in Section \ref{sec:rel3} and compared among them.
\subsection{Accuracy}
Before delving into why the model produced such exact output, it is necessary to establish an accurate prediction model. As a consequence, inherently or post-hoc interpretable models must be assessed on their capacity to attain high accuracy when given the same set of datasets in order to compare their performance objectively. As indicated before, the XCM architecture has shown its performance and that of the MTEX-CNN on classification tasks utilizing the UEA Archive of diverse MTS datasets \cite{bagnall2018uea}. The comparisons are done using model accuracy as the assessment measure, and a table of model accuracy reports is then generated for each of the experimental models over all datasets in the UEA archive. Additionally, a critical difference chart is constructed to illustrate the performance of each model more intuitively by aligning them along a line marked with the difference level from the reported accuracy table.

\subsubsection{Datasets}

The UEA multivariate time series classification archive \cite{bagnall2018uea} has 30 datasets that span across six categories, including Human Activity Recognition, Motion Classification, ECG Classification, EEG/MEG Classification, and Audio Spectra Classification. As was the case with other sources of datasets, this was a collaborative effort between academics at the University of California, Riverside (UCR) and the University of East Anglia (UEA). All the time series within one dataset has the same length, and no missing values or infinity values occur. 

\subsubsection{Accuracy metrics}

After training on the aforementioned datasets, each model is assessed using the following accuracy score. 
\begin{equation}
Accuracy = \frac{TP + TN}{TP + FP + TN + FN},
\end{equation}
where TP, FP, TN and FN are abbreviations for True Positive, False Positive, True Negative, and False Negative, respectively. The nominator (TP + TN) denotes the number of predictions that are equal to the actual class, while FP and FN indicate the number of predictions that are not equal to the true class. 

Following that, the average score is utilized to generate a Critical Difference Diagram that depicts any statistically significant difference between the architectures. It is created using the Dunn's Test, a nonparametric technique for determining which means are more significant than the others. The Dunn's test establishes a null hypothesis, in which no difference exists between groups, and an alternative hypothesis, in which a difference exists between groups.

\subsection{Interpretability}

Despite developing the interpretable CNN-based architecture and using Grad-CAM for interpretation, MTEX-CNN and XCM evaluate their explainability only using a \% Average Drop metric and a human comprehensibility test on trivial simulation data. The majority of significant testing of CAM-based explanations are undertaken inside the CAM-based technique. Score-CAM \cite{wang2020score} assesses their approach on the most comprehensive collection of trials merged from the other method, which is appropriate given the method's recent publication. The assessments include metrics for evaluating faithfulness, such as Increase of Confidence, percent Average Drop, percent Average Increase, and Insertion/Deletion curves; metrics for evaluating localization, such as the energy-based pointing game; and finally, a sanity check by examining the change in the visualization map when a random set of feature maps is used. By contrast, attention-based RNN techniques are primarily concerned with studying the attention's spatiotemporal properties. As with the percent Average Drop, multiple approaches (DA-RNN, DSTP, and STAM) perform the ablation experiment by viewing the difference between the unattended and original multivariate time series data. Thus, a single set of interpretability evaluation trials should collect all the dispersed testing across the methodologies in order to serve as a baseline for comparing the effectiveness of each interpretation produced by each approach in the area of MTS classification. This includes human visual inspection, fidelity of the explanations to the model parameters, spatiotemporality of the explanation, and causality of the explanation in relation to the model parameters. 

Because the assessment is based on the interpretation of a given model prediction, before extracting an explanation for the model of choice, the model must be trained with at least chance-like accuracy. Since the explanation for attention-based models is fundamental to the model parameters, explanations can be retrieved exclusively for the output class, but this is not the case with CAM-based approaches. Additionally, to ensure that interpretations become clearer, the interpretation process should be conducted on a single dataset with a maximum of three component time series. Thus, \textit{UWaveGestureLibrary} is chosen for interterpretability evaluation.

\subsubsection{Faithfulness}

The evaluations that belong to this class attempt to justify whether the features that one explaining mechanism figures out are consistent with the outcomes of the model or not. These consist of two sub-classes, namely Average Drop/Average Increase and Deletion/Insertion AUC score. 

Average Drop and Average Increase are included together as a measure because they both test the same feature of an explanation's faithfulness to the model parameters, but the Deletion/Insertion AUC score analyzes a different aspect. According to \cite{chattopadhay2018grad}, given $Y^c_i$ as the prediction of class $c$ on image $i$ and $O^c_i$ as the prediction of class $c$ on image $i$ but masked by the interpretation map, the Average Drop is defined as in equation \ref{eq:AverageDrop} \cite{wang2020score}, whereas the Average Increase, also called the Increase of Confidence, is computed using equation \ref{eq:AverageIncrease}. \\
\begin{equation}
AverageDrop(\%) = \frac{1}{N} \sum^N_{i=1}\frac{max(0, Y^c_i - O^c_i)}{Y^c_i}*100
\label{eq:AverageDrop}
\end{equation}
\begin{equation}
AverageIncrease(\%) = \sum^N_{i=1}\frac{Sign(Y^c_i < O^c_i)}{N}*100
\label{eq:AverageIncrease}
\end{equation}
where $Sign$ is the function that converts boolean values to their binary counterparts of $0$ and $1$. As the names of these approaches imply, an interpretability method performs well when the Average Drop percentage lowers and the Average Increase percentage grows.

The Deletion and Insertion AUC score measures are intended to be used in conjunction with the Average Drop and Average Increase measurements. The deletion metric reflects a decrease in the predicted class's probability when more and more crucial pixels are removed from the generated saliency map. A steep decline in the graph between the deletion percentage and the prediction score, equivalent to a low-lying area under the curve (AUC), indicates a plausible explanation. On the other hand, the insertion measure captures the increase in likelihood associated with the addition of additional relevant pixels, with a larger AUC implying a more complete explanation \cite{wang2020score}.

\subsubsection{Causality}

When doing a causality test, it is common practice to assign random numbers to the causes and see how they behave in response to those numbers. Using randomization, we may display different pieces of evidence that point to a causal link. This is accomplished by randomly assigning each feature vector one by one in a cascade method until all of the feature vectors of the input data are completely randomized. It is also necessary to randomize the time dimension up to the final time point, as previously stated. Each piece of randomized data is then put into the interpretable models in order to extract its interpretation matrix, which is then connected with the original explanations to see how far it is deviating from them. If the interpretation does not alter from the original one, it is possible that causal ties will be severed since the interpretation will be invariant with respect to the data input. The correlation values between the randomized input explanations and their root explanation without randomization serve as a quantitative evaluation of the tests. It is possible that the correlations produced using the same interpretable technique diverged, as shown by a drop in correlation factors, during the course of all cascading stages, and that this may be used as a shred of evidence for the existence of causal relationships. The Chi-square Goodness-of-fit hypothesis testing procedure is used to determine whether the divergence is significant enough to cause a difference between the yielded interpretations from the randomization and the initial interpretation map, with the null hypothesis being that there is no difference between the two interpretations yielded by the randomization. In other words, the null hypothesis is that the correlation between the original explanation and the observed data is 1, while the alternative hypothesis is that the correlation is the opposite. According to the following definition, the Chi-square is
\begin{equation}
\chi^2 = \sum \frac{(O_i - E_i)^2}{E_i},
\label{eq:chisquare}
\end{equation}
where $O_i$ denotes the observed value, which in this case is the correlation of interpretations obtained by input randomization, and $E_i$ is the predicted value, which is $1$, indicating a perfect match to the original interpretation map. All of the data in the \textit{UWaveGestureLibrary}'s test set is evaluated in this manner, where the quantity of data minus one is the degree of freedom for the Chi-square test.

\subsubsection{Spatiotemporality}

The spatiotemporality of a multivariate time series specifies and distributes the relevance weights for each time step of each feature vector. The metric for determining the explanation map's spatiotemporality is as straightforward as ensuring both temporality and spatiality. In other words, the interpretation map must be adaptable in both time and space. For example, when $N$ feature vectors and m time steps are used in a multivariate time series, it ensures spatiality when the summation of interpretation map values along the time axis for each feature does not equal $\frac{1}{N}$. Similarly, it would fulfill temporality if the total of the map along the feature axis for each time step $t$ did not equal $\frac{1}{T}$. If one of these properties fails, the related property fails as well, which results in the spatiotemporality as a whole failing.
These criteria are expressed mathematically in equations \ref{eq:Spatiality} and \ref{eq:Temporality}.
\begin{equation}
\sum_j X_{nj} \neq \frac{1}{N} \quad \forall{n} \in \{0, ..., N-1 \}
\label{eq:Spatiality}
\end{equation}
\begin{equation}
\sum_i X_{it} \neq \frac{1}{t} \quad \forall{t} \in \{0, ..., T-1 \}
\label{eq:Temporality}
\end{equation}

\subsection{Experiment settings}

Due to the fact that XCM and TSEM allow for parameter adjustment to a fraction of the data length via time windows calculated in multiple layers of the architecture, it is either unfair to other architectures with a fixed-parameter setting or it makes the models themselves so large that the computational capabilities cannot handle the training and may result in overfitting to the training data. As a consequence, the number of their architectural parameters varies between datasets, and the time frame is set to the proportion that results in absolute values no more than 500. This is not the case with MTEX-CNN, since the number of parameters is fixed. The other attention-based RNN architectures employ an encoder-decoder structure, which will be set to $512$ units for both the encoder and decoder modules. 

This assessment portion was entirely implemented using Google Colab Pro and the Paperspace Gradient platform. Google Colab Pro Platform includes a version of Jupyter Lab with 24 GB of RAM and a P100 graphics processing unit (GPU) card with 16 GB of VRAM for inference machine learning models that need a CUDA environment. Similarly, Paperspace's Gradient platform enables users to connect with the Jupyter Notebook environment through a 30 GB RAM configuration with additional GPU possibilities up to V100, and evaluations are conducted using a P6000 card with 24 GB VRAM. 

\subsection{Experiment results}

The experiment results are presented by two folds: accuracy and interpretability, which has been further broken down into four sections, namely human visual evaluation, faithfulness, causality and spatiotemporality.
\subsubsection{Accuracy}
As indicated before, this part evaluates $10$ interpretable models on $30$ datasets. Table \ref{tab:ueaacc1} summarizes the findings. According to the table,
	\begin{table}[ht!]
    \caption{Accuracy evaluation of the interpretable models for each dataset with TSEM (DSTP is shorthand for DSTP-RNN, DSTP-p is shorthand for DSTP-RNN-parallel, GeoMAN-l and GeoMAN-g are shorthands for GeoMAN-local and GeoMAN-global respectively)}
    \centering
    \small\addtolength{\tabcolsep}{-6pt}
    \label{tab:ueaacc1}
    \begin{adjustbox}{width=1.3\textwidth}
    \begin{tabular}{||c@{\hskip 0.1in}c@{\hskip 0.1in}c@{\hskip 0.1in}c@{\hskip 0.1in}c@{\hskip 0.1in}c@{\hskip 0.1in}c@{\hskip 0.1in}c@{\hskip 0.1in}c@{\hskip 0.1in}c@{\hskip 0.1in}c@{\hskip 0.1in}c||} 
    \hline
    {\textbf{Datasets}} & \textbf{MTEX} &  {\textbf{XCM}} & {\textbf{TSEM}} & \textbf{DA} &  {\textbf{RETAIN}} & \textbf{DSTP} &  {\textbf{DSTP}} &  {\textbf{GeoMAN}} & \textbf{GeoMAN} & \textbf{GeoMAN} &  {\textbf{STAM}} \\ [0.5ex] 
    & \textbf{-CNN} & & & \textbf{-RNN} &  & \textbf{-p} &  &  & \textbf{-g} & \textbf{-l} &  \\ [0.5ex] 
    \hline\hline
    ArticularyWordRecognition &  0.837 & 0.6 & 0.557 & 0.893 & 0.903 & 0.846 & 0.85 & 0.92 & 0.906 & 0.923 & \textbf{0.97} \\ 
    \hline
    AtrialFibrillation &  0.333 & 0.4667 & 0.4667 & 0.4 & 0.4 & 0.4 & \textbf{0.6} & 0.4 & 0.4667 & 0.333 & 0.533 \\ 
    \hline
    BasicMotions &  0.9 & 0.75 & 0.925 & 0.9 & 0.85 & 0.8 & 0.875 & \textbf{0.95} & \textbf{0.95} & 0.925 & 0.675 \\ 
    \hline
    CharacterTrajectories &  \textbf{0.065} & 0.06 & 0.06 & 0.06 & 0.06 & 0.06 & 0.06 & 0.06 & 0.06 & 0.06 & 0.06 \\     \hline
    Cricket &  0.083 & 0.583 & 0.722 & 0.208 & 0.208 & 0.153 & 0.194 & 0.194 & 0.208 & 0.194 & \textbf{0.75} \\ 
    \hline
    DuckDuckGeese &  0.2 & \textbf{0.54} &0.4& 0.42 & 0.32 & 0.28 & 0.26 & 0.36 & 0.4 & 0.38 & 0.42 \\ 
    \hline
    EigenWorms &  0.42 & \textbf{0.428} & 0.42 & 0.42 & 0.42 & 0.42 & 0.42 & 0.42 & 0.42 & 0.42 & 0.412 \\ 
    \hline
    Epilepsy &  0.601 & 0.804 & \textbf{0.891} & 0.348 & 0.312 & 0.384 & 0.384 & 0.333 & 0.341 & 0.326 & 0.565 \\ 
    \hline
    EthanolConcentration &  0.251 & 0.32 &\textbf{0.395}& 0.32 & 0.346 & 0.297 & 0.357 & 0.327 & 0.312 & 0.323 & 0.308 \\  
    \hline
    ERing&  0.619 & 0.696 &\textbf{0.844}& 0.47 & 0.756 & 0.478 & 0.441 & 0.426 & 0.463 & 0.459 & 0.692 \\  
    \hline
    FaceDetection &  0.5 & 0.5 &0.513& 0.5 & 0.545 & 0.518 & 0.515 & 0.517 & 0.517 & 0.63 & \textbf{0.65} \\ 
    \hline
    FingerMovements &  0.51 & 0.54 & 0.53& 0.6 & 0.6 & 0.53 & \textbf{0.62} & 0.61 & 0.53 & 0.52 & 0.56 \\  
    \hline
    HandMovementDirection &  0.405 & \textbf{0.54} &0.514& 0.446 & 0.46 & 0.487 & 0.378 & 0.527 & 0.473 & 0.392 & 0.527 \\ 
    \hline
    Handwriting &  0.051 & 0.095 &\textbf{0.117}& 0.051 & 0.061 & 0.055 & 0.051 & 0.051 & 0.037 & 0.051 & 0.099 \\ 
    \hline
    Heartbeat &  \textbf{0.8727} & 0.771 &0.746& 0.722 & 0.756 & 0.722 & 0.722 & 0.722 & 0.722 & 0.727 & 0.756 \\  
    \hline
    JapaneseVowels &  \textbf{0.238} & \textbf{0.238} &0.084& 0.084 & 0.084 & 0.084 & 0.084 & 0.084 & 0.084 & 0.084 & 0.084 \\ 
    \hline
    Libras &  0.067 & 0.411 &0.372& 0.206 & 0.372 & 0.201 & 0.228 & 0.233 & 0.272 & 0.172 & \textbf{0.589} \\ 
    \hline
    LSST &  0.315 & 0.155 &0.315& 0.315 & 0.315 & 0.315 & 0.315 & 0.315 & 0.315 & 0.315 & \textbf{0.316} \\  
    \hline
    InsectWingbeat &  \textbf{0.01} & \textbf{0.01} &\textbf{0.01}& \textbf{0.01} & \textbf{0.01} & \textbf{0.01} & \textbf{0.01} & \textbf{0.01} & \textbf{0.01} & \textbf{0.01} & \textbf{0.01} \\  
    \hline
    MotorImagery &  0.5 & 0.5 &0.5& 0.54 & 0.51 & 0.56 & 0.52 & \textbf{0.63} & 0.59 & 0.56 & 0.56 \\ 
    \hline
    NATOPS &  0.8 & \textbf{0.844} &0.833& 0.344 & 0.661 & 0.233 & 0.228 & 0.25 & 0.333 & 0.522 & 0.767 \\ 
    \hline
    PEMS-SF &  0.67 & 0.549 & 0.544& 0.636 & \textbf{0.775} & 0.168 & 0.145 & 0.162 & 0.671 & 0.688 & 0.746 \\ 
    \hline
    PenDigits &  - & 0.721 &0.686& 0.323 & 0.746 & 0.112 & 0.11 & 0.331 & 0.35 & 0.384 & \textbf{0.888} \\  
    \hline
    Phoneme &  0.026 & \textbf{0.07} & 0.058&0.066 & 0.049 & 0.037 & 0.059 & 0.05 & 0.068 & 0.042 & 0.06 \\ 
    \hline
    RacketSports &  0.533 & 0.75 &\textbf{0.77}& 0.283 & 0.447 & 0.283 & 0.283 & 0.29 & 0.29 & 0.336 & 0.441 \\  
    \hline
    SelfRegulationSCP1 &  0.502 & 0.747 &0.836& 0.604 & \textbf{0.898} & 0.58 & 0.604 & 0.58 & 0.563 & 0.87 & 0.877 \\  
    \hline
    SelfRegulationSCP2 &  0.502 & 0.517 &\textbf{0.756}& 0.583 & 0.533 & 0.561 & 0.539 & 0.567 & 0.544 & 0.561 & 0.556 \\ 
    \hline
    SpokenArabicDigits &  \textbf{0.01} & \textbf{0.01} &\textbf{0.01}& \textbf{0.01} & \textbf{0.01} & \textbf{0.01} & \textbf{0.01} & \textbf{0.01} & \textbf{0.01} & \textbf{0.01} & \textbf{0.01} \\ 
    \hline
    StandWalkJump &  0.333 & 0.467 &0.467& 0.4 & 0.333 & 0.467 & 0.333 & 0.467 & 0.467 & \textbf{0.533} & \textbf{0.533} \\ 
    \hline
    UWaveGestureLibrary&  0.725 & 0.813 &\textbf{0.831}& 0.497 & 0.781 & 0.406 & 0.497 & 0.466 & 0.375 & 0.444 & 0.813 \\  
    \hline
    \hline
    Average Rank & 6.5 & 3.7 & \textbf{3.5}& 5.1 & 4.3 & 6.1 & 6.1 & 5.0 & 5.1 & 5.2 & \textbf{3.2} \\
    \hline
    Wins/Ties &  5 & 8 & \textbf{9} & 2 & 4 & 2 & 4 & 4 & 3 & 3 & \textbf{9} \\[1ex]
    \hline
    \end{tabular}
    \end{adjustbox}
    \end{table}
TSEM, XCM and STAM models have significantly different average rankings and win/tie times when compared to the others. In which, TSEM has significantly higher accuracy towards datasets having long sequences such as Cricket and SelfRegulationSCP2 which have a length of 1197 and 1152 time steps, respectively \cite{bagnall2018uea}. RETAIN also performs well in comparison to the other approaches in terms of the average rank. While MTEX-CNN has the lowest average rank, it has the highest number of wins/ties among the methods except from TSEM, XCM and STAM, indicating that this approach is unstable and not ideal for all types of datasets. In comparison, despite the fact that DA-RNN, GeoMAN-local, and GeoMAN-global acquire techniques with the fewest wins/ties, they have a solid average rank. Both DSTP-RNN and DSTP-RNN-parallel produce the same average rank; however, DSTP-wins/ties RNN's times outnumber those of DSTP-RNN-parallel and this indicates that its behavior is consistent with the corresponding research on DSTP-RNN, which indicates that DSTP-RNN can remember a longer sequence than DSTP-RNN-parallel. Otherwise, the performance of DSTP-RNN and DSTP-RNN-parallel is not superior to the others, as their report shown in the regression issue, but it is the poorest of all the approaches. \\
\begin{figure}[h!]
\centering
\includegraphics[width=0.99\textwidth]{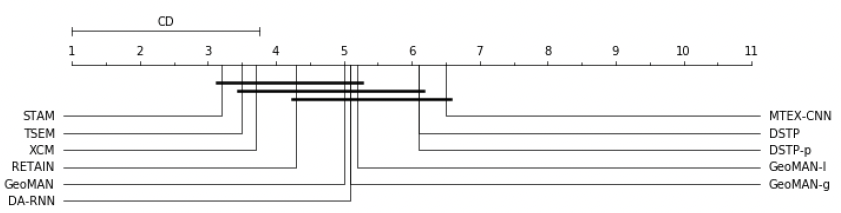}
\caption{The Critical Difference plot of the MTS classifiers on the UEA datasets with alpha equals to
0.05.}
\label{fig:CCD}
\end{figure}
\\

While comparing average rankings and wins/ties score might help determine a model's quality, there may be discrepancies between these two measurements. Average rank is  included in the Critical  Difference Diagram to provide further insight. The statistical test shown in Fig. \ref{fig:CCD} was developed by Bonferroni-Dunn with an alpha of 0.05 and 30 datasets examined, which corresponds to the total number of datasets in the UEA collection. Fig. \ref{fig:CCD} suggests that STAM, TSEM and XCM are the top-3 methods in terms of accuracy performance, and they are in the same group with RETAIN, DA-RNN and all the variants of GeoMAN posing a significant difference to the remaining three methods. 
\subsubsection{Qualitative evaluation}
Regarding CNN-based architectures that are analyzed in this research, ten different CAM-based methods are evaluated for the explainability of the models. They are CAM \cite{zhou2016learning}, Grad-CAM \cite{selvaraju2017grad}, Grad-CAM++ \cite{chattopadhay2018grad}, Smooth Grad-CAM++ \cite{omeiza2019smooth}, XGrad-CAM \cite{fu2020axiom}, Ablation-CAM \cite{ramaswamy2020ablation}, Score-CAM \cite{wang2020score}, Integrated Score-CAM \cite{naidu2020cam}, Activation-Smoothed Score-CAM \cite{wang2020ss} and Input-Smoothed Score-CAM \cite{wang2020ss}. However, only CAM, Grad-CAM++, and XGrad-CAM are examined in this part due to the supremacy of these three methods. Then, they are visually compared with the attention representation vectors of the attention-based RNN architectures with in a given context. 

\begin{figure}[h!]
\centering
\includegraphics[width=0.9\textwidth]{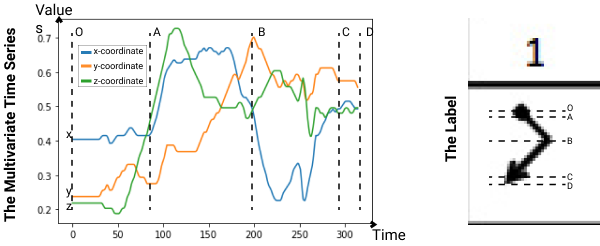}
\caption{A \textit{UWaveGestureLibrary} class 1 instance with semantic segmentation.}
\label{fig:uwave}
\end{figure}
Here, an example from the \textit{UWaveGestureLibrary} dataset (shown in Fig. \ref{fig:uwave}) illustrates a right turn downhill after a straight walk onwards. As previously mentioned, if there is no sanity check provided by human understanding, the explanations for a multivariate time series would be subtle. In other words, every element in the multivariate time series must be unambiguous about what it is supposed to be doing. To provide an example, the input data in Fig. \ref{fig:uwave} depicts a multivariate time series from the \textit{UWaveGestureLibrary} dataset that corresponds to the three axes of an accelerometer that measure a staged action, which is designated by the label 1 in the category list. Knowing what each time series component represents, for example, knowing that the blue line represents the x-axis values of the acceleration sensor, it becomes clearer why they oscillate in certain patterns, as shown here by the blue line's oscillation following the x-axis variation of the steep right turn as labeled. In particular, the action intervals for class number 1 are denoted with great precision. Indeed, the action would differ from person to person and from time to time, but in general, it can be divided into five states and four stages based on the change in acceleration in the x-axis values of the sensor as reported by the uphill and downhill patterns of the blue line and the uphill and downhill patterns of the green line. As a result, the five states are denoted by the letters O, A, B, C and D, which correspond to the resting state, the beginning state, the switching state, the halting state, and the terminating state, in that order. The four phases are represented by the letters OA, AB, BC and CD, which represent the temporary stage, the first direction running stage, the second direction running stage, and the concluding stage, respectively, in the graphical representation. It is necessary to distinguish between transitory and concluding phases since an action is neither initiated or terminated immediately after a signal has been initiated or terminated by a device. All of the phases and stages are highlighted in both the multivariate time series and its label in order to demonstrate the sensible interconnections between them. So an interpretation map is more interesting if it highlights the critical points that are located between stages A and B; otherwise, it would be meaningless if it stressed the transient stage OA or the ending stage CD because, logically, one model should not choose class 1 over other classes simply because of its longer transient stage, for example.
\begin{figure}[h!]
\centering
\includegraphics[width=1.2\textwidth]{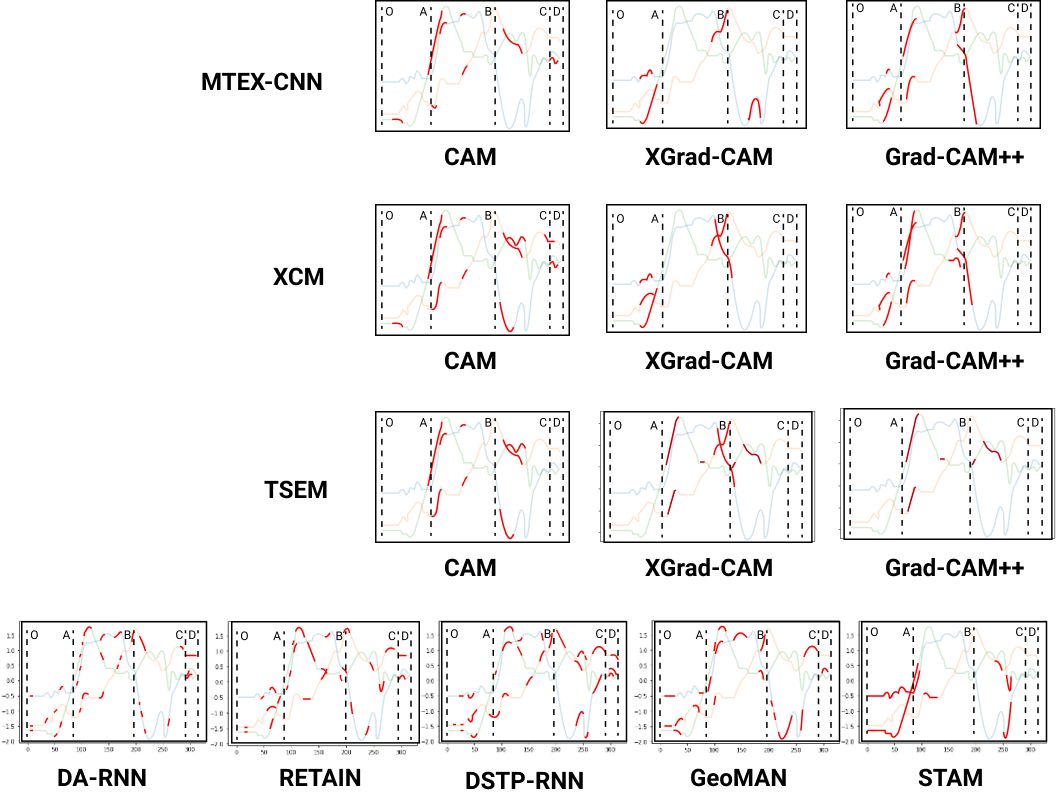}
\caption{Explanation of some explainable models for a \textit{UWaveGestureLibrary} class 1 instance. MTEX-CNN, XCM and TSEM are post-hoc explained by CAM, XGrad-CAM and Grad-CAM++. Explanations for attention-based RNN methods are their spatiotemporal attention vectors. The red lines show the highest activated regions of the input for class 1 as their predictions.}
\label{fig:expatvis}
\end{figure}

In general, according to Fig. \ref{fig:expatvis}, post-hoc explainability approaches based on CAM have yielded more continuous interpretation maps for XCM than explainability methods based on attention-based RNN models. It appears that the difference is due to the different nature of using CNN and RNN for extracting the learned features, with CNN being able to provide a local explanation specific to the input instance, whereas RNN is believed to yield a global explanation independent of the specific input instance in question. For the sake of this discussion, the explanation map produced by the attention in recurrent networks is designated for the category to which the input instance is categorized as a whole, and this category is represented by a node in the instance. Unlike RNN, CNN does not have the capacity to memorize; the highlighted areas are simply those parts of the input instance that are aroused when the CNN encounters a label. As a result, the interpretation based on CAM is strongly reliant on the input signal.

\subsubsection{Faithfulness}

As a contrastive pair, the two assessment metrics in each test are shown on a two-dimensional diagram, together with the correctness of each model, which is represented by the size of the circle representing the coordinates. When high-accuracy interpretable models are compared against low-accuracy interpretable models, the accuracy might reveal how the explanations are impacted.
\begin{figure}[h!]
\centering
\includegraphics[width=1.2\textwidth]{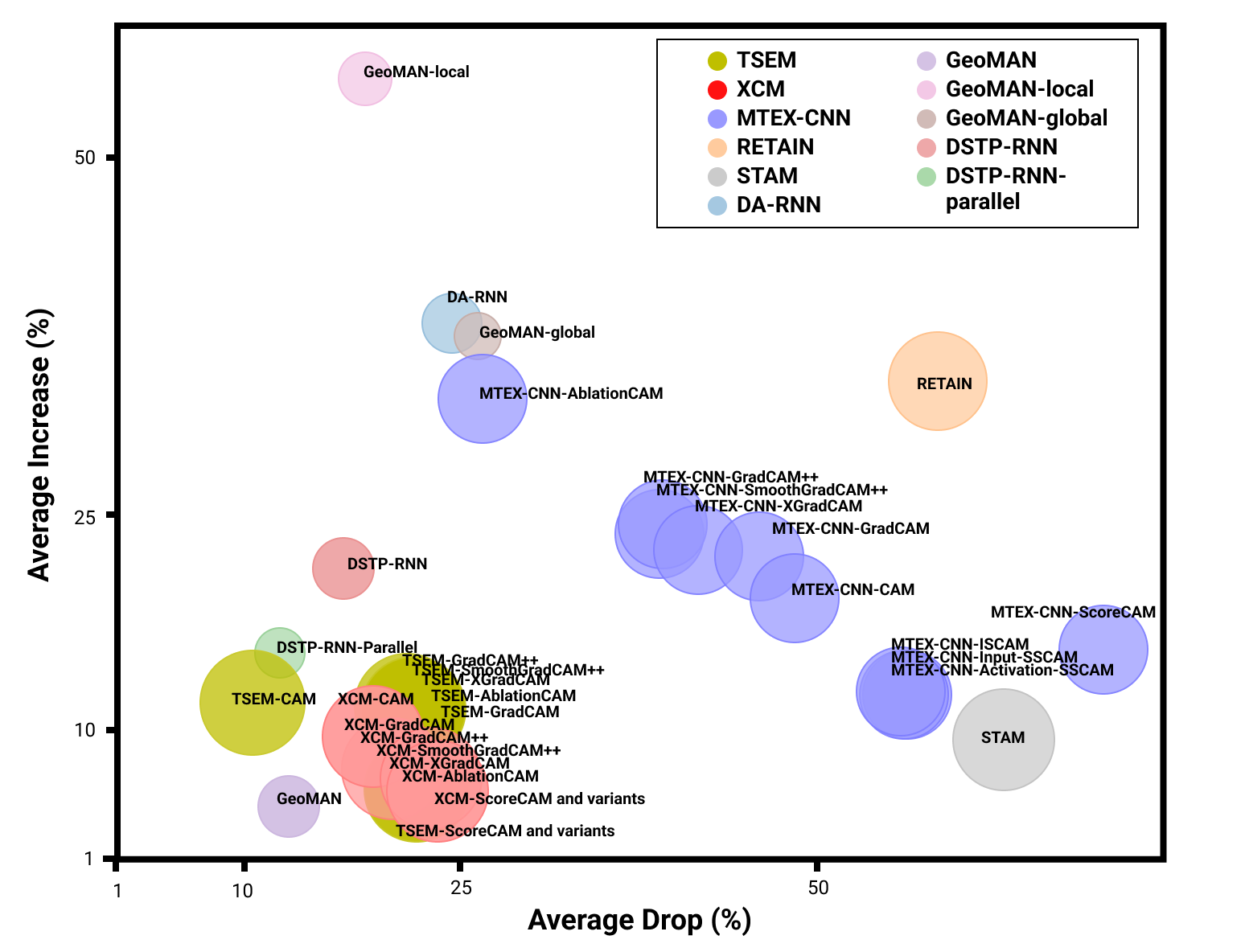}
\caption{The Average Drop - Average Increase Diagram for the \textit{UWaveGestureLibrary} dataset. Accuracy
is illustrated as proportional to the size of the circles. (The lower the Average Drop is, the more faithful
the method get, as contrary to the Average Increase).}
\label{fig:tsemadai}
\end{figure}
The Average Drop and Average Increase metrics are displayed as percentages ranging from $0$ to $100$, and each point is represented by two coordinates that correspond to the average drop and average increase metrics. Because the Average Drop and Average Increase are inversely proportional, it is predicted that all of the points would follow a trend parallel to the line $y = x$. Considering that $x$ represents the Average Drop value and y represents the Average Increase value, the right bottom is the lowest performance zone, while the left top is the highest performance sector in this equation. Taking into account the wide range of accuracy at each stage, it is rather simple to determine which technique provides the most accurate explanation for the model's conclusion. Figure \ref{fig:tsemadai} depicts the distributions of interpretation techniques in terms of their association with the Average Drop and the Average Increase in percentages. All of the points are color-coded, with the red and yellow colors denoting the spectrum of CAM-based interpretations for the two different XCMs (XCM and TSEM), respectively. The blue hue represents MTEX-CNN, while the remaining colored spots represent the visualization for the attention-based approaches (which are not shown here). In general, the farther to the right the figure is moved, the poorer the implied performance becomes. As previously mentioned, all of the worst approaches are clustered together at the bottom right, where the Average Increase is at its lowest value and the Average Drop is at its greatest value, which is a significant difference. While both MTEX-CNN and STAM have great accuracy (as shown by the size of the circles), the collection of interpretations for MTEX-CNN and the attention vector for STAM have the lowest fidelity to their judgments.
\begin{figure}[h!]
\centering
\includegraphics[width=1.2\textwidth]{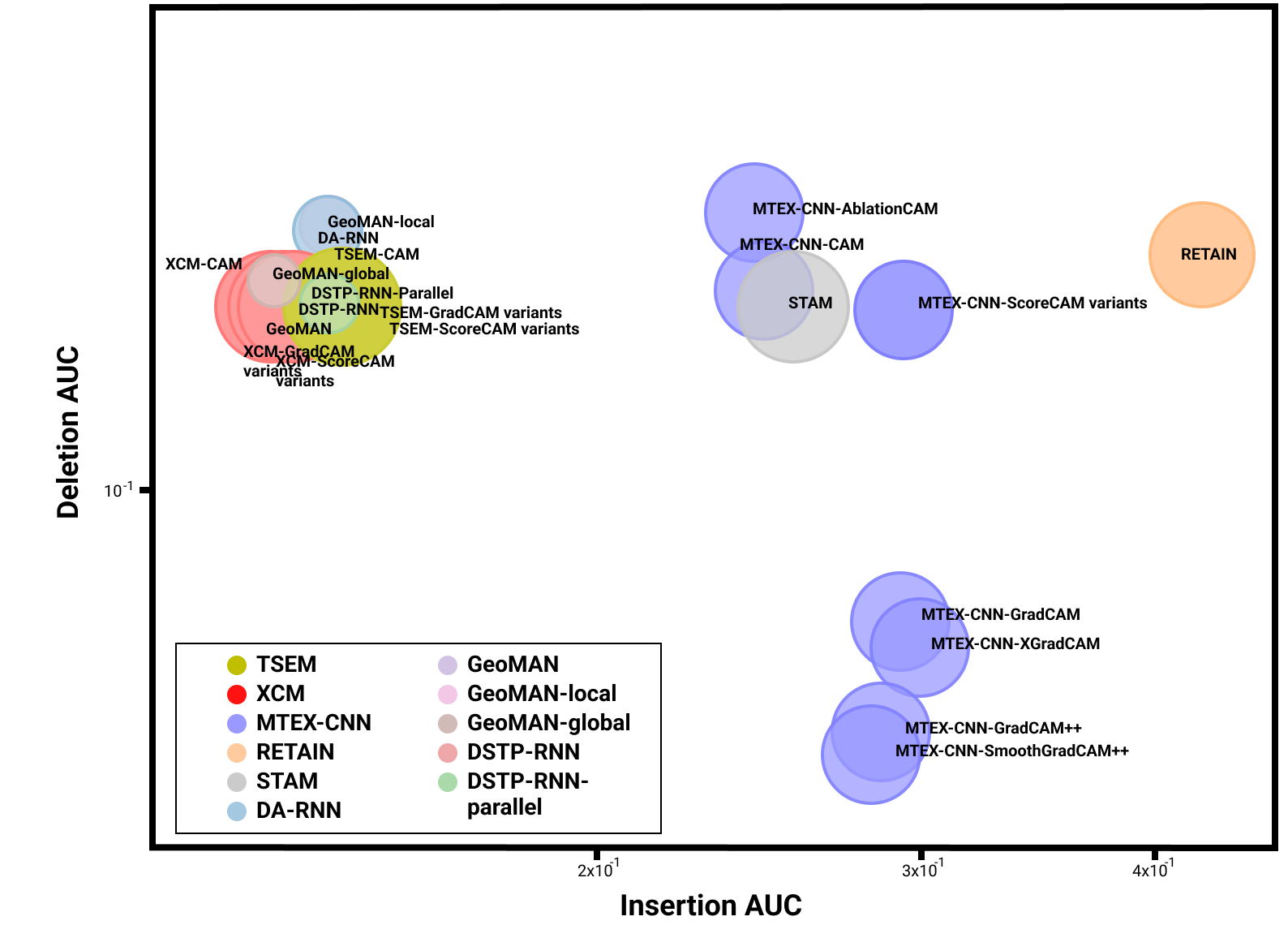}
\caption{The Deletion/Insertion AUC Diagram for the \textit{UWaveGestureLibrary} dataset. Accuracy is
illustrated as proportional to the size of the circles. The diagram is shown in log-scale to magnify the distance between circles for a clearer demonstrative purpose. (The lower the Deletion AUC score is, the more faithful the method get, as contrary to the Insertion AUC score).}
\label{fig:tsemdaia}
\end{figure}
When only essential data points are considered in the input, the change in the models' prediction score is gradually introduced to an empty sequence or is gradually removed from the original input data, and this is reflected through the Insertion and Deletion curves, respectively. The area under each curve serves as a measure, providing information about how quickly the curve is moving. It is anticipated that the area under the curve (AUC) of a Deletion Curve be as minimal as feasible, showing the rapid suppression of the model accuracy when the most relevant data points are beginning to be eliminated. Contrary to this, the AUC of an Insertion curve should be as substantial as feasible, which suggests that accuracy increases as soon as the initial most essential data points are added. Figure \ref{fig:tsemdaia} depicts the relationship between Deletion AUC values and Insertion AUC values for a given sample size. Overall, there are no evident patterns in the AUC values or the accuracy of any approach when seen as a whole. When compared to their Insertion AUC values, the majority of the techniques are grouped in accordance with nearly equal Deletion AUC around $0.125$, however the Grad-CAM collective approaches for MTEX-CNN stand out due to their remarkable Deletion AUC scores varied about $0.07$. Furthermore, not only do they have low Deletion AUC scores, but they also have a high Insertion AUC score, which is around $0.3$. This is where STAM and the remaining CAM-based methods for MTEX-CNN are located, with Deletion AUC scores that are almost doubled when compared to the Grad-CAM collective for MTEX-CNN. Similarly, the RETAIN interpretation has the highest Insertion AUC score, which is approximately $0.5$, which is four times higher than the XCM interpretations and four times higher than the rest of the attention-based techniques' interpretations, with the exception of STAM. While XCM CAM-based explanations have nearly the same Insertion AUC score as GeoMAN, GeoMAN-local, GeoMAN-global, DA-RNN, DSTPRNN, and DSTP-RNN-parallel, they have the lowest Deletion AUC scores among the attention maps of GeoMAN, GeoMAN-local, GeoMAN-global, DA-RNN, DSTPRNN, and DSTP-RNN-parallel. 

It is hoped that the fidelity of the CAM-based explanations for TSEM will be at least as good as that of the XCM architecture, having been modified and corrected from the XCM design. Indeed, as seen in Fig. \ref{fig:tsemadai} and \ref{fig:tsemdaia}, the cluster for TSEM interpretation in both diagrams is distributed in a manner that is virtually identical to that of XCM. The TSEM, on the other hand, performs somewhat better in terms of two metrics: average increase and Insertion AUC (area under the curve). This implies that TSEM interpretations pay more attention to data points that are more meaningful in the multivariate time sequence. Most notably, the original CAM approach extracts an explanation map for TSEM with the smallest Average Drop when compared to the other methods tested.

\subsubsection{Causality}

When attempting to reason about an effect in relation to a cause, the significance of causality cannot be overstated. Specifically, the effect corresponds to the explanation map that corresponds to its cause, which is the input data, and is connected to the cause and effect by means of a model that acts as a proxy between the cause and effect. For example, in contrast to the regression connection between a model's input and output, explanation maps are produced as a result of a mix of inputs, model parameters, and outputs.
\begin{figure}[h!]
\centering
\includegraphics[width=1.2\textwidth]{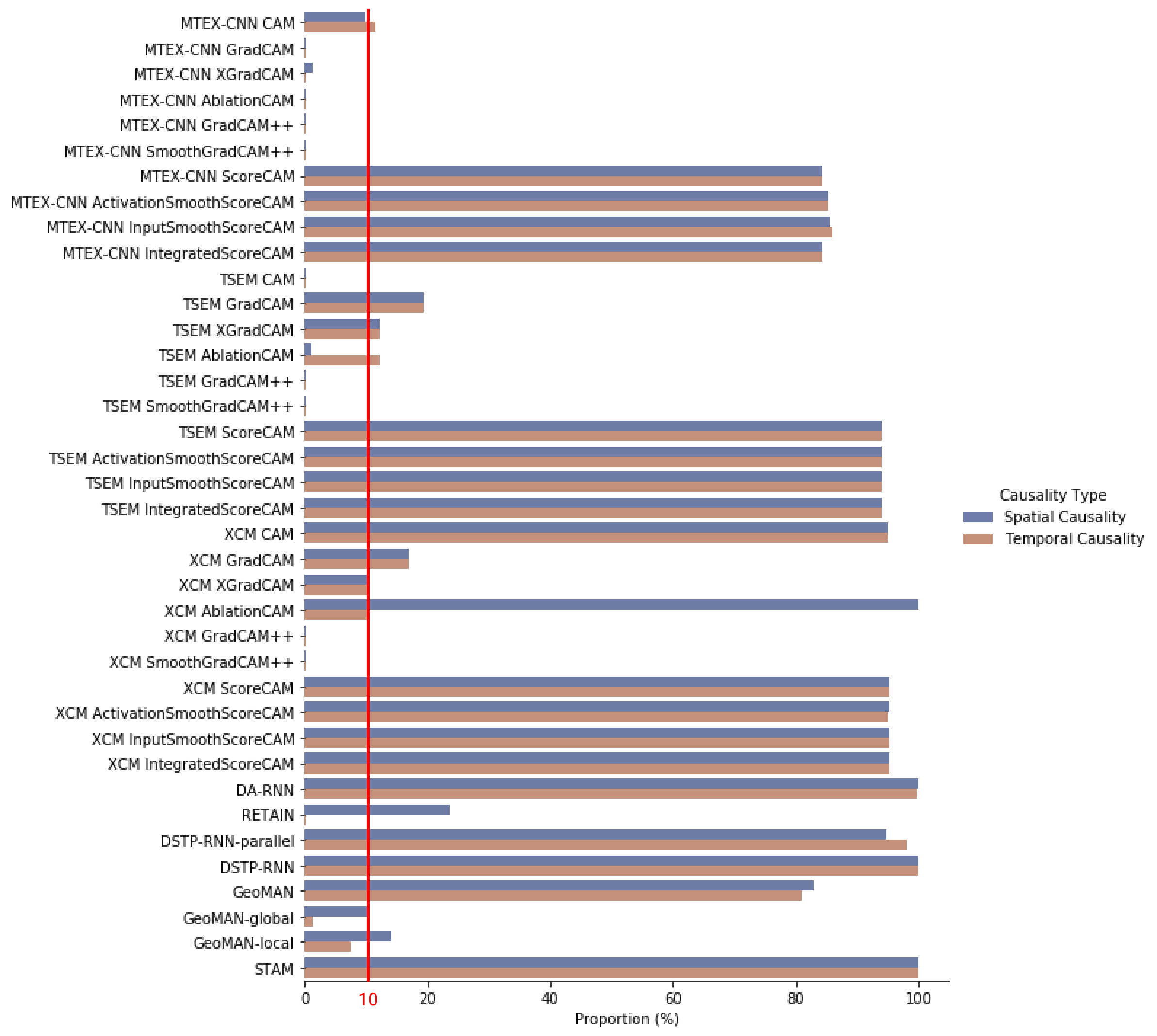}
\caption{The Bar Chart of non-causal proportion of \textit{UWaveGestureLibrary} test set inferred by TSEM CAM-based explanations vs the other interpretable methods. The lower proportion is, the better causation level a method gets and it must be below 10\% to be considered (illustrated by the red line) pass the causality test.}
\label{fig:causality}
\end{figure}

While the model parameters and the output are maintained constant in this assessment, randomization is applied to two axes of the multivariate time series input. According to Fig. \ref{fig:causality}, the CAM-based explanations for TSEM with 320 occurrences in the \textit{UWaveGestureLibrary} test set exhibit similar patterns to the XCM explanations. Specifically, none of the Score-CAM variants nor the Ablation-CAM variants pass the causality test. The distinction between TSEM and MTEX-CNN and XCM is that it renders the original CAM approach causal, which does not happen with MTEX-CNN or XCM. Additionally, the XGrad-explanations CAM's for TSEM are non-causal. Although the Ablation-CAM explanation technique fails in both XCM and TSEM, the failure is more pronounced in TSEM when the proportion of temporal non-causal data points surpasses the 10\% threshold. In general, only three approaches satisfy the causality criteria for TSEM: Grad-CAM++, Smooth Grad-CAM++, and the original CAM. This is considered a good performance, because according to Fig. \ref{fig:causality}, almost $70\%$ the number of models do not retain causality. 

\subsubsection{Spatiotemporality}

This assertion is made clearly in Equations \ref{eq:Spatiality} and \ref{eq:Temporality}, which relate to the spatiality and temporality tests, respectively. If both of these equations apply to an interpretation map, it is deemed to possess the spatiotemporal quality. Because the numbers in an explanation map do not add up to 1, they must be normalized before applying the criterion equations. This is done by dividing each value by the total of the whole map. All CAM-based method interpretations in XCM, TSEM and MTEX-CNN, as well as the attention-based interpretation, pass this set of tests. Because no negative instances are provided, the findings for each approach are omitted. 

\section{Conclusion and outlook}

After a thorough analysis of the currently available interpretable methods for MTS classification, the Temporally weighted Spatiotemporal Explainable network for Multivariate Time Series Classification, or TSEM for short, is developed on the basis of the successful XCM in order to address some of the XCM's shortcomings. Specifically, XCM does not permit concurrent extraction of spatial and temporal explanations due to their separation into two parallel branches. Simultaneously, TSEM reweights the spatial data obtained in the first branch using the temporal information learnt from the recurrent layer in the second parallel branch. This is regarded to be a more productive method than XCM in terms of extracting real temporality from data rather than pseudo-temporality from the correlation of time-varying values in location. This also lends credence to an explanation including maps of the relative significance of temporal and spatial features. As a result, it is expected to provide a more compact and exact map of interpretation. Indeed, TSEM outperforms XCM in terms of accuracy across over 30 datasets in the UEA archive and in terms of explainability in the \textit{UWaveGestureLibrary}. 

This study focuses only on model-specific interpretable approaches and makes no comparisons to model-independent methods. Thus, it would be interesting if TSEM could be analyzed with these methods using the same evaluation set of interpretability metrics. In any other case, TSEM would use concept embedding in its future work to encode tangible aspects into knowledge about a concept. After that, it would incorporate neuro-symbolic technique in order to provide a more solid explanation towards its prediction. In addition to this, causal inference has to be considered in order to get rid of any false connection in the logits and the feature maps. This would help strengthen the actual explanation and get rid of any confounding variables that may be present. 

%
% ---- Bibliography ----
%
\bibliographystyle{tsem/bibtex/spmpsci}
\bibliography{tsem}

\end{document}